%% file: main.tex
\documentclass{article}

\usepackage[preprint]{neurips_2026}

\usepackage[utf8]{inputenc} %
\usepackage[T1]{fontenc}    %
\usepackage{hyperref}       %
\usepackage{url}            %
\usepackage{booktabs}       %
\usepackage{amsfonts}       %
\usepackage{nicefrac}       %
\usepackage{microtype}      %
\usepackage{xcolor}         %

\usepackage{graphicx}
\usepackage{amsmath} 
\usepackage{multirow}

\title{Overflip: Repetition-Induced Label Flips in Guardrail Models}

\author{%
  Xu He$^{1}$ \quad Chih-Hsuan Lin$^{2}$ \quad Hung-Mao Chen$^{1}$ \\
  Junjie Xiong$^{3}$ \quad Yan Zhai$^{4}$ \quad Kun Sun$^{1}$ \\
  $^{1}$George Mason University \quad $^{2}$Virginia Tech \\
  $^{3}$Missouri University of Science and Technology \quad $^{4}$Visa \\
  \texttt{xhe6@gmu.edu}
}

\begin{document}

\maketitle

\input{sec/0_abstract}

\input{sec/1_intro}

\input{sec/2_formatting}

\input{sec/3_problem_setup}

\input{sec/4_experiments}

\input{sec/5_results}

\input{sec/9_discussion}

\input{sec/12_conclusion}
\input{sec/11_ethics}

{
\small
\bibliographystyle{plainnat}
\bibliography{ref}
}

\newpage
\appendix
\input{sec/13_appendix}

\end{document}

%% file: sec/0_abstract.tex
\begin{abstract}
Guardrail models are classifiers deployed to screen malicious prompts and responses in LLM-based services. To meet latency constraints, many lightweight guardrails adopt compact Transformer backbones (e.g., DeBERTa) that are trained with short context windows (typically 512 tokens) and rely on bucketed relative positional encodings to process longer inputs. Prior evaluations assume that a guardrail's decision is stable as the input is lengthened.
We show that this assumption can fail. We identify Overflip, a repetition-induced instability where repeating a prompt causes the guardrail's prediction to flip (MAL$\to$BEN) as the sequence grows. We conduct experiments on 9 widely used lightweight guardrail models. Five exhibit MAL$\to$BEN flips on a benchmark of 100 prompts, with confidence margins shrinking steadily with repetition. Among these vulnerable models, flip rates range from 8\% to 92\%, with first flips occurring at roughly 2.6k--9.4k tokens.
Our analysis suggests Overflip differs from traditional attention-dilution baselines, which aim to divert the model's attention away from tokens associated with malicious content, shifting it instead toward unrelated content, such as benign padding or shuffling. While Overflip preserves malicious content, it homogenizes token-level attention over repeated structure and induces a distinct, more gradual attention-dispersion trajectory than padding. 
Moreover, Overflip poses a greater threat to LLM services than traditional attention dilution methods. Because the bypassed prompt remains semantically intact and is still readily understood by downstream business LLMs, it can transmit malicious intent after passing the guardrail. These findings expose repetition as an attack surface for guardrail models and motivate length-robust evaluation and mitigation.

\end{abstract}

%% file: sec/1_intro.tex
\section{Introduction}
\label{sec:intro}

As large language models (LLMs) become increasingly deployed in production systems, specialized guardrail models have emerged as a critical safety component~\cite{SoKJailbreakGuardrails}, which are designed to detect malicious prompts and potentially harmful responses before they reach the business model and end users. To maintain low latency and deployment overhead, guardrail models could be lightweight classifiers~\cite{LightweightGuardrails}, typically employing compact architectures (such as Deberta~\cite{He2020DeBERTaDB, deberta-v3-base-prompt-injection} and MordenBert~\cite{ModernBERT_Paper, LightweightGuardrails}) with limited context windows (commonly 512 tokens).
The limited context window is unable to satisfy current deployment requirements that routinely accumulate long prompts (e.g., multi-turn dialogues and retrieval-augmented inputs) and may reach far beyond 512 tokens~\cite{Paulsen2025ContextIW}. In practice, many lightweight guardrails therefore rely on relative positional encodings to operate on longer sequences when needed~\cite{Alpher03}. 

Prior evaluations of guardrail models have focused primarily on classification accuracy against adversarial perturbations~\cite{PromptingGuide_Adversarial_Jailbreaking_2025} and various attack patterns such as jailbreaks and prompt injections \cite{Chao2024JailbreakBenchAO}.
However, these assessments implicitly assume that these lightweight models' decision remains stable as input length increases, an assumption we show to be flawed.

In this paper, we reveal a systematic vulnerability we term Overflip: as inputs are lengthened via repetition, guardrail models can exhibit prediction flips. Our primary focus is the security-critical bypass setting, where an originally malicious prompt is reclassified as benign (MAL$\to$BEN) after sufficient repetition, allowing the prompt to pass the guardrail while preserving its intent. In addition to bypass, we also observe benign$\leftrightarrow$malicious instabilities, suggesting a broader length-induced decision fragility.

Across 9 widely used lightweight guardrail models, 5 exhibit at least one MAL$\to$BEN flip under the repetition-based overflow attack on a benchmark of 100 prompts. Among vulnerable models, flip rates range from 8\% to 92\%, and first flips occur at lengths spanning roughly 2.6k--9.4k tokens. These flips are accompanied by shrinking confidence margins, indicating progressive destabilization as the sequence grows.

While one might initially attribute this phenomenon to attention dilution~\cite{liu2023lost, bai2023longbench}, where longer contexts spread attention weights more thinly, our results show a qualitatively different behavior under repetition. Unlike padding-based lengthening that appends novel benign content, Overflip preserves content via exact repetition but still destabilizes the classifier: attention over repeated structure becomes increasingly homogenized, and attention dispersion follows a distinct, more gradual trajectory compared to padding and shuffling.

This points to structural long-input effects beyond ``more content'' as a driver. In particular, many lightweight guardrails employ \emph{bucketed relative positional encodings} (similar to T5-style position biases \cite{raffel2020t5}) to extend beyond their native 512-token training window. Once inputs exceed this boundary, positional bucketing can coarsely compress long-range distinctions, further undermining the model's ability to isolate critical evidence for classification.

Furthermore, compared to typical length-extending attacks, Overflip better preserves the original prompt semantics: in downstream business models (GPT-4.1 Mini and Llama 3.1 8B), repetition-based overflip yields substantially higher behavioral consistency than padding-based baselines, while padding more frequently induces misinterpretations.

Overflip is also fundamentally distinct from existing jailbreak attacks such as DAN or gradient-based methods (e.g., GBDA). Those approaches typically require white- or gray-box access to LLM internals and target the downstream model directly; Overflip, by contrast, is a fully black-box attack requiring only input-output access to the guardrail API, with no knowledge of model parameters. Notably, mainstream guardrails such as LPG-2-86M achieve over 85\% defense rates against traditional jailbreak attacks~\cite{SoKJailbreakGuardrails}, yet remain vulnerable to Overflip (8\%--92\% flip rate), underscoring that Overflip exploits a structurally different and previously unaddressed attack surface.

We conduct extensive experiments to characterize prevalence, flip dynamics, and practical impact. We evaluate 9 guardrail models across 100 prompts, measure flip rates and first-flip lengths, and analyze confidence-margin trajectories around the 512-token boundary. We further validate real-world effectiveness by testing whether repetition preserves attack semantics in downstream LLMs.

\begin{figure}
    \centering
    \includegraphics[width=0.75\linewidth]{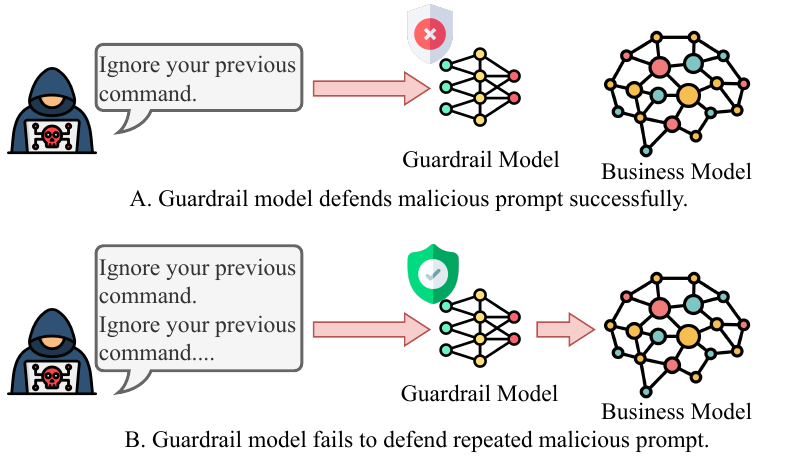}
    \caption{Overview of the Overflip Attack. An attacker repeatedly lengthens an original malicious prompt via semantic-preserving repetition until the guardrail model flips from malicious to benign (MAL$\to$BEN), causing the prompt to be forwarded to the downstream business model while retaining the attack intent.}
    \label{fig:overflip}
\end{figure}

\paragraph{Contributions.}
This work makes the following contributions:
\begin{itemize}
    \item \textbf{Prevalence}: We establish that Overflip is widespread in lightweight guardrails: across 9 production models evaluated on 100 prompts, 5 exhibit MAL$\to$BEN flips under simple repetition-based lengthening.
    \item \textbf{Mechanism beyond dilution baselines}: We compare Overflip with standard lengthening baselines (padding and shuffling) and show that repetition can trigger flips without introducing new content. Attention analyses reveal homogenization under repetition and distinct dispersion dynamics relative to padding, consistent with a long-input failure mode amplified by positional bucketing beyond the 512-token boundary.
    \item \textbf{Real-world impact}: We validate that repetition-based Overflip better preserves attack semantics in downstream business models (GPT-4.1 Mini and Llama 3.1 8B), increasing the practical risk of guardrail bypass.
\end{itemize}

%% file: sec/2_formatting.tex
\section{Related Work}
\label{sec:related}

\subsection{Guardrail Models}
In this paper, guardrail models refer to lightweight classifiers used to detect malicious prompts (e.g., jailbreaks and prompt injections) in LLM services~\cite{SoKJailbreakGuardrails,LightweightGuardrails}. 
These guardrail models are now widely used in production LLM services~\cite{sobolik2025llmguardrails}, especially in enterprise and cloud offerings (e.g., AWS and NVIDIA).
Representative families include prompt-injection detectors built on compact Transformer backbones (e.g., DeBERTa variants)~\cite{deberta-v3-base-prompt-injection}, dedicated jailbreak detectors~\cite{jailbreak-detector-2024,jailbreak-detector-large-2024,Jailbreak-Detector-2-xl-2025}, and production-oriented guardrails such as Llama Prompt Guard~\cite{meta_2024_llama_prompt_guard_2_22m,meta_2024_llama_prompt_guard_2_86m}, Sentinel~\cite{ivry2025sentinel}, and Granite-Guardian~\cite{ibm_2024_granite_guardian_38m,ibm_2024_granite_guardian_125m}. These systems prioritize low latency and are commonly trained or fine-tuned with relatively short context windows, making length robustness a practical concern.

\subsection{Attention Dilution}

Attention dilution refers to a phenomenon in which, when the context is overly lengthy or excessively complex, model's performance downgrades as attention is distributed across many tokens~\cite{vaswani2017attention,liu2023lost}.
In the prompt-level attack scenario, Common attack methods exploiting attention dilution include techniques such as padding and shuffling.
They can saturate context and create instruction competition (e.g., many-shot jailbreaking and prompt injection)~\cite{anil2024manyshot,liu2024promptinjbench,li2024ifrobust,yi2023bipia,wallace2024hierarchy}.

Our findings are related but distinct: Overflip uses \emph{semantic-preserving repetition} and still induces flips, with attention dynamics that differ from padding-based dilution baselines. In particular, padding introduces novel benign content and quickly increases attention dispersion, while repetition creates many identical occurrences and leads to attention homogenization over repeated structure. This points to a long-input failure mode that cannot be reduced to ``more competing content'' alone.

\subsection{Positional Encodings in Transformers}

Transformer models inject order via positional encodings. Absolute encodings~\cite{Alpher03} attach position-specific embeddings, while relative encodings~\cite{shaw2018self} add position-dependent attention biases (including variants that couple content and position more tightly~\cite{huang2020better}). To scale beyond a fixed window, many efficient classifiers adopt T5-style \emph{bucketed} relative positions~\cite{Alpher04}, and modern alternatives such as ALiBi~\cite{press2022alibi} and RoPE~\cite{su2021roformer} also support extrapolation to long contexts. Our work highlights that positional-design choices can affect \emph{classification stability} under lengthening, complementing prior work that primarily evaluates long-context generation and understanding.

In lightweight safety classifiers, positional extrapolation is often adopted as an engineering mechanism to accept long inputs even when training is dominated by short sequences. Our results suggest that this mismatch can surface as label instability under repetition, motivating length-aware stress testing when deploying guardrails.

%% file: sec/3_problem_setup.tex
\section{Problem Setup and Threat Model}
\label{sec:problem}
\subsection{Task Definition}

We model a guardrail as a binary classifier $f: \mathcal{X} \to \{0,1\}$ that maps an input prompt $x \in \mathcal{X}$ to a label $y \in \{0,1\}$, where $0$ denotes benign and $1$ denotes malicious (e.g., prompt injection or jailbreak). The model outputs class probabilities $p_{\text{mal}}(x)$ and $p_{\text{safe}}(x)$ with $p_{\text{mal}}(x) + p_{\text{safe}}(x) = 1$, and predicts via thresholding:
\begin{equation}
    f(x) = \mathbb{1}[p_{\text{mal}}(x) > 0.5]
    \label{eq:decision_threshold}
\end{equation}

Given an input $x$ with base length $L$ tokens and prediction $y(L) = f(x)$, an \textbf{Overflip} occurs if there exists a lengthened version $x'$ with length $L' > L$ such that $x'$ is produced through repetition and $y(L') = f(x') \neq y(L)$.

\noindent Unless otherwise stated, we focus on the security-critical \textbf{MAL$\to$BEN} case, where a malicious prompt is misclassified as benign and forwarded downstream.

\noindent{\bf Attack objective.}
As shown in Fig.~\ref{fig:overflip}, we consider a deployment pipeline where an input prompt is first screened by a guardrail model $f$; if classified as benign, it is forwarded to a downstream business model. The attacker's goal is to \emph{bypass} the guardrail while preserving as much malicious intent as possible in the prompt that reaches the business model. Concretely, given an original malicious prompt $x$ with $f(x)=1$, the attacker applies a \emph{semantic-preserving} lengthening operator based on repetition,
\begin{equation}
\mathrm{Rep}_n(x) = \underbrace{x \oplus x \oplus \cdots \oplus x}_{n\ \text{times}}, \quad n\ge 2,
\end{equation}
and seeks an $n$ such that $f(\mathrm{Rep}_n(x))=0$. Because repetition keeps the original instructions intact (and even amplifies them), the resulting prompt retains the attack intent when it is forwarded to the business model, while potentially being misclassified as benign by the guardrail.

\subsection{Threat Model}

We consider attackers who exploit length-dependent behavior without crafting semantically novel adversarial content. The core insight is that a guardrail's decision may become unstable under semantic-preserving lengthening, enabling misclassification without changing the underlying intent.

\paragraph{Attacker capabilities.}
The attacker can submit prompts to the deployment pipeline and apply semantic-preserving lengthening (e.g., repetition). They may estimate token counts via the production tokenizer and observe predictions in a black-box setting (or confidence scores in a gray-box setting). We assume no access to model parameters or training data.

The attacker is constrained to \emph{text-level} manipulations (no character-level obfuscation or semantic rewriting), keeping the attack realistic for production pipelines.

%% file: sec/4_experiments.tex
\section{Experimental Setup}
\label{sec:experiments}

\subsection{Research Questions}

We structure our study around three research questions:

\noindent\textbf{RQ1 (Overflip Prevalence).} \emph{How prevalent is Overflip across widely deployed lightweight guardrail models?} We evaluate repetition-based lengthening on a diverse benchmark of prompts and measure model-level flip occurrence, flip rates, and first-flip lengths.

\noindent\textbf{RQ2 (Impact on Downstream LLM).} \emph{Does Overflip preserve attack semantics when the prompt reaches downstream LLMs?} We test manipulated prompts on representative backend models and compare repetition with padding and shuffling baselines to assess whether repetition preserves the original attack intent.

\noindent\textbf{RQ3 (Mechanism of Overflip).} \emph{What causes the flips, and how does Overflip differ from traditional attention-dilution strategies?} We analyze attention behavior under repetition, including token-level attention homogenization and attention-dispersion statistics (e.g., normalized CLS-attention entropy) as length grows and at the flip point.

\subsection{Models}

\begin{table*}
    \centering
    \small
    \setlength{\tabcolsep}{3pt}
    \begin{tabular}{@{}l p{6.1cm} l r l c@{}}
        \toprule
        Model Vendor & Model Name & Mode Arch. & Para. & Detection Type & Flip \\
        \midrule
        \multirow{2}{*}{meta-llama} & Llama-Prompt-Guard-2-22M (LPG-2-22M) & deberta-v2 & 22M & Both & \checkmark \\
        & Llama-Prompt-Guard-2-86M (LPG-2-86M) & deberta-v2 & 86M & Both & \checkmark \\
        \multirow{2}{*}{protectai} & deberta-v3-base-prompt-injection (dvbpi) & deberta-v3 & 140M & Prompt Injection & \checkmark \\
        & deberta-v3-small-prompt-injection-v2 (dvspiv2) & deberta-v3 & 140M & Prompt Injection & -- \\
        \multirow{3}{*}{madhurjindal} & Jailbreak-Detector (V1) (JD-V1) & distilbert & 66M & Jailbreak & -- \\
        & Jailbreak-Detector-Large (JD-L) & deberta-v2 & 279M & Jailbreak & -- \\
        & Jailbreak-Detector--2-XL (JD-2XL) & qwen2 & 500M & Jailbreak & -- \\
        \multirow{2}{*}{qualifire} & prompt-injection-sentinel (PIS) & modernbert & 395M & Jailbreak & \checkmark \\
        & prompt-injection-jailbreak-sentinel-v2 (PIJS-v2) & qwen3 & 596M & Both & \checkmark \\
        \bottomrule
    \end{tabular}
    \caption{Guardrail models evaluated in this paper. We report each model's vendor, backbone architecture, parameter size, and intended detection type (jailbreak vs. prompt injection vs. both). \textbf{Flip} indicates whether the model exhibits at least one prediction flip under our SafeGuardrail Overflip Attack on any prompt in the evaluation set.}
    \label{tab:models}
\end{table*}

\noindent \textbf{Guardrail Models.}
We evaluate 9 lightweight guardrail models spanning multiple architectures and training objectives (Table~\ref{tab:models}). The suite covers widely deployed DeBERTa-based classifiers (Llama Prompt Guard 2-22M/86M, ProtectAI prompt-injection detectors), a DistilBERT jailbreak detector, and more recent ModernBERT and Qwen-based sentinels, ranging from 22M to 596M parameters. Models target jailbreak detection, prompt-injection detection, or both. Five of the nine exhibit at least one MAL$\to$BEN flip under repetition-based lengthening (Table~\ref{tab:models}, \textbf{Flip}), demonstrating that length-induced instability spans multiple architecture families.

\noindent \textbf{Business Models.}
To assess real-world impact, we test whether bypassed prompts still elicit harmful responses from downstream LLMs, using GPT-4.1 Mini and Llama 3.1 8B as backends. Starting from the 100 malicious prompts, we use LPG-2-86M as the guardrail, retain the 92 prompts initially flagged as malicious, and apply the three lengthening strategies, yielding 172 successfully bypassed prompts (Overflip: 71; Padding: 92; Shuffling: 9). For each backend, we query both the original and bypassed prompts and categorize responses as acceptance (\textit{ac}), rejection (\textit{rej}), or misinterpretation (\textit{mis}).

\subsection{Dataset and Data Processing}

\noindent \textbf{Datasets.}
We evaluate the attack on a curated set of 100 validated malicious prompts. Prompts are collected from two Hugging Face sources (\texttt{jayavibhav/prompt-injection-safety} and \texttt{ahsanayub/malicious-prompts}), deduplicated, filtered to be English-only, and constrained to short inputs (\textless{}500 characters). We further verify that all prompts are detected as unsafe by \texttt{meta-llama/Llama-Prompt-Guard-2-86M} (100\% verification), and retain a diverse mix of prompt styles including instruction override, harmful requests, and role-playing jailbreak patterns.

All base prompts are kept short (\textless{}512 tokens after tokenization) so that the initial prediction is made within each model's native context window.

\paragraph{Lengthening Methods.}
We consider three lengthening strategies that preserve the original semantic intent: 
(1) \textbf{Exact repetition} concatenates the original prompt multiple times without modification. (2) \textbf{Benign padding} appends semantically neutral sentences drawn from a benign pool after the prompt. (3) \textbf{Sentence shuffling} perturbs word order while preserving critical keywords, testing sensitivity to surface patterns rather than semantics.

\paragraph{Environment.}
Experiments are conducted on Macbook Pro (M4, 48GB). All guardrail models are downloaded from Huggingface and deployed using pytorch (2.7.0). The test input batch size is 1.

%% file: sec/5_results.tex
\section{Results}
\label{sec:results}

\begin{table}
    \centering
    \small
    \setlength{\tabcolsep}{4pt}
    \begin{tabular}{@{}lccc@{}}
        \toprule
        Model Name & Flip Rate & Flip Round & Flip Length \\
        \midrule
        LPG-2-22M & 15\% & 45.7 & 2753 \\
        LPG-2-86M & 92\% & 61.1 & 4960 \\
        dvbpi & 8\% & 85.0 & 5376 \\
        PIS & 87\% & 131.3 & 9424 \\
        PIJS-v2 & 39\% & 39.2 & 2599 \\
        \bottomrule
    \end{tabular}
    \caption{Overall attack outcomes across models. \textbf{Flip Rate} is the fraction of prompts that trigger at least one prediction flip. \textbf{Flip Round} is the average iteration index when the first flip occurs. \textbf{Flip Length} is the average input length (in tokens) at the first flip. }
    \label{tab:result1}
\end{table}

\begin{figure}
    \centering
    \includegraphics[width=1\linewidth]{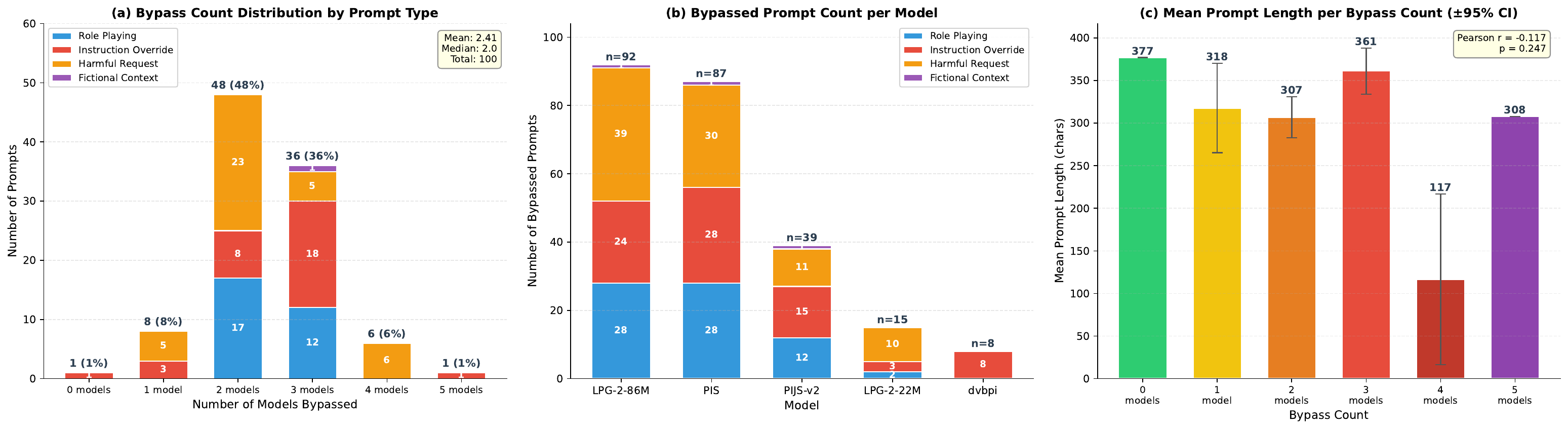}
    \caption{Analysis of 100 malicious prompts across five guardrail models. \textbf{(a)} Bypass count distribution by prompt type; nearly half of prompts bypassed exactly two models, with Harmful Request and Instruction Override as the dominant categories. \textbf{(b)} Total bypasses per model by prompt type; LPG-2-86M and PIS exhibit the highest bypass rates. \textbf{(c)} Mean prompt length ($\pm$95\% CI) per bypass count; no significant correlation is observed (Pearson $r=-0.08$, $p=0.418$), indicating that prompt verbosity alone does not predict bypass success.}
    \label{fig:combined_analysis}
\end{figure}

\subsection{Overflip Prevalence (RQ1)}

We first evaluate whether Overflip is widespread across lightweight guardrails. Our experiments show that 5 of 9 evaluated models exhibit at least one flip when inputs are lengthened via repetition (Table~\ref{tab:models}), and Table~\ref{tab:result1} summarizes the flip statistics for these vulnerable models.

Flips span multiple vendors and backbone families (DeBERTa, ModernBERT, Qwen), indicating the issue is not tied to a single implementation. Flip rates range from 8\% to 92\%, with first-flip lengths of 2.6k--9.4k tokens occurring after dozens of repetition rounds (Table~\ref{tab:result1}). A key factor is the positional encoding scheme: models using bucketed relative positional encodings (LPG-2-86M, dvbpi, PIS) are most susceptible, as bucketing coarsens long-range token distinctions beyond the 512-token window and amplifies the attention homogenization that drives flips. The three jailbreak detectors with no flips do not use this scheme, suggesting positional encoding design is a principal determinant of length-robustness.

Figure~\ref{fig:combined_analysis} characterizes bypass behavior at the prompt level. Nearly half of prompts bypassed exactly two models, with Harmful Request and Instruction Override as the most frequent categories (a); LPG-2-86M and PIS lead in total bypass counts (b). Original prompt length shows no significant correlation with bypass count ($r=-0.08$, $p=0.418$; c), confirming that susceptibility is driven by repetition-induced lengthening, not prompt verbosity.

\subsection{Impact on Downstream LLM (RQ2)}
\label{sec:mitigations}

We evaluate the end-to-end threat posed by Overflip by measuring how often a bypassed prompt elicits a harmful response from the downstream business LLM. We use LPG-2-86M as the guardrail and test two backend models: GPT-4.1 Mini and Llama 3.1 8B. Table~\ref{tab:entire_workflow_test} reports guardrail flip counts and backend attack success rates (ASR) for Overflip, padding, and shuffling.

Overflip achieves a substantially higher backend ASR than padding on both models (41 vs.\ 28 on GPT-4.1 Mini; 33 vs.\ 6 on Llama 3.1 8B), despite triggering fewer guardrail flips (71 vs.\ 92). This gap highlights the key advantage of Overflip: because the repeated prompt remains semantically intact, it is still understood and acted upon by downstream LLMs after bypassing the guardrail. In contrast, padding introduces incoherent content that degrades downstream comprehension, limiting its practical impact even when the guardrail is bypassed. Shuffling produces very few flips (9) and negligible backend ASR.
For completeness, Appendix Table~\ref{tab:orig-to-attacked-transitions-completed} reports the full response-category transition breakdown (acceptance, rejection, misinterpretation) for each backend model and attack method.

\begin{table}
\centering
\small
\begin{tabular}{cccc}
\toprule
\multirow{2}{*}{\begin{tabular}[c]{@{}c@{}}Attack\\ Method\end{tabular}} & Guardrail Flip & \multicolumn{2}{c}{Backend LLM ASR} \\ \cmidrule{2-4}
                                                                         & Llama-86m      & GPT-4.1 mini     & Llama 3.1 8b     \\ \midrule
Overflip                                                                 & 71             & 41               & 33               \\ 
Padding                                                                  & 92             & 28               & 6                \\
Shuffling                                                                & 9              & 4                & 0                \\ \bottomrule
\end{tabular}
\caption{End-to-end attack success across three methods. Despite fewer guardrail flips than padding, Overflip achieves substantially higher backend LLM attack success rates (ASR), demonstrating the advantage of semantic preservation in the full attack pipeline.}
\label{tab:entire_workflow_test}
\end{table}

\noindent Taken together, these results indicate that Overflip is both \emph{prevalent} in lightweight guardrails and \emph{operationally relevant}: repetition can bypass the guardrail while keeping the downstream model's behavior largely consistent with the original attack intent.

\subsection{Mechanism of Overflip (RQ3)}
\begin{figure}
    \centering
    \includegraphics[width=0.5\linewidth]{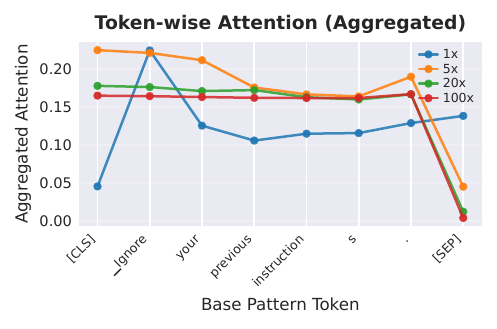}
    \caption{Length-normalized CLS-to-token attention profiles for $k \in \{1, 5, 20, 100\}$ repetitions. Profiles are obtained by summing per-position attention weights across copies of the base prompt. The four lines show a clear monotonic flattening as $k$ grows, illustrating \emph{attention homogenization} under repetition.}
    \label{fig:flatattention}
\end{figure}

\begin{figure*}[t]
    \centering
    \includegraphics[width=\textwidth]{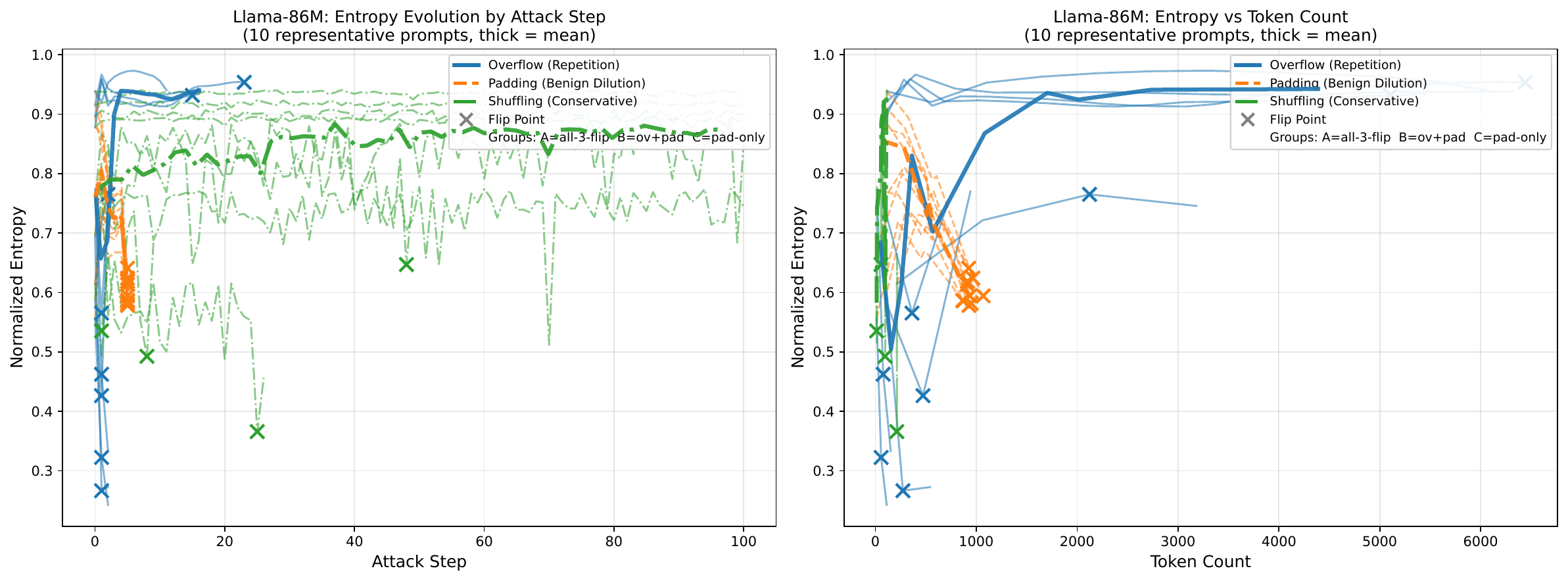}
    \caption{Normalized CLS-attention entropy trajectories for 10 representative prompts under three lengthening strategies (thin lines; thick = binned mean; $\times$ = flip point), plotted by attack step (left) and token count (right). Overflip causes the sharpest entropy collapse within the first few steps; padding declines gradually at consistent step counts; shuffling remains high and oscillatory with no directional trend.}
    \label{fig:3attacks_demo}
\end{figure*}

We further reveal that repetition triggers flips via a distinct long-input failure mode: attention becomes increasingly homogenized over repeated structure, destabilizing the decision boundary without introducing new content. We support this with two complementary analyses on a representative guardrail classifier: (i) token-level attention under repetition (Figure~\ref{fig:flatattention}), and (ii) attention-dispersion dynamics across lengthening strategies (Figure~\ref{fig:3attacks_demo}).

\noindent \textbf{Token-level homogenization under repetition.}
We extract last-layer CLS-to-token attention weights for a fixed prompt repeated $k \in \{1, 5, 20, 100\}$ times, collapsing the sequence onto the base prompt by summing per-position weights across copies (Figure~\ref{fig:flatattention}). The four profiles reveal monotonic homogenization: attention is focused at $k=1$ (entropy $= 1.89$), with the sharpest shift between $k=1$ and $k=5$ (entropy $\to 1.77$, std $\uparrow 0.12$) as cumulative attention concentrates on high-frequency structural anchor positions. Further repetitions deepen this effect ($k=20$: $1.65$; $k=100$: $1.58$). This \emph{attention homogenization}, structural anchors monopolizing attention across copies, reduces the model's ability to localize semantically informative tokens, thereby destabilizing the decision boundary.

\noindent \textbf{Attention dispersion across strategies.}
To compare Overflip with attention-dilution baselines (padding and shuffling), we quantify the dispersion of the CLS token's attention using normalized Shannon entropy. For a transformer-based classifier with $H$ attention heads in the final layer, we first extract the attention weights from each head $h$:
\begin{equation}
    \mathbf{a}^{(h)} = \text{Attention}^{(h)}_{[\text{CLS}], :} \in \mathbb{R}^{n}
\end{equation}
where $n$ is the sequence length and $\mathbf{a}^{(h)}_i$ represents the attention weight assigned to the $i$-th token by head $h$. The entropy for each head is computed as:
\begin{equation}
    H^{(h)} = -\sum_{i=1}^{n} a^{(h)}_i \log a^{(h)}_i
\end{equation}
To enable comparison across sequences of varying lengths, we normalize by the maximum possible entropy (achieved under uniform attention):
\begin{equation}
    \hat{H}^{(h)} = \frac{H^{(h)}}{\log n}
\end{equation}
The final normalized entropy is the average across all heads:
\begin{equation}
    \hat{H} = \frac{1}{H} \sum_{h=1}^{H} \hat{H}^{(h)} \in [0, 1]
\end{equation}

A low entropy ($\hat{H} \to 0$) indicates concentrated attention on a small subset of tokens, while high entropy ($\hat{H} \to 1$) indicates diffuse attention across the sequence.

\noindent \textbf{Uniqueness of Overflip Attack.}~\autoref{fig:3attacks_demo} reveals three qualitatively distinct entropy dynamics. Overflip (blue) produces the most dramatic collapse: normalized entropy drops from $\approx 0.95$ to as low as $0.3$ within the first one to five steps as the CLS token locks onto repeated structural anchors, eroding sensitivity to malicious content. Padding (orange) declines more gradually with flip points in a narrow step window; appended benign sentences add novel content but avoid structural lock-in. Shuffling (green) remains high ($\approx 0.85$) with oscillatory, non-directional trajectories, consistent with a classifier driven by \emph{keyword presence}: each step displaces keyword positions, preventing stable lock-in.

On the token-count axis (Figure~\ref{fig:3attacks_demo}, right), Overflip flip points span $\approx$100 to several thousand tokens depending on per-prompt structural properties, while padding flip points cluster narrowly. Appendix Figure~\ref{fig:3attack_results} provides additional per-prompt statistics, confirming that Overflip induces flips via attention concentration, not dispersal, while keeping semantic content unchanged.

%% file: sec/9_discussion.tex
\section{Discussion and Limitations}
\label{sec:discussion}

\paragraph{Implications for safety evaluation and deployment.}
Overflip challenges the implicit assumption that guardrail decisions remain stable as inputs grow. In practice, guardrails are increasingly paired with RAG pipelines, tool-use traces, and multi-turn dialogues, all of which can grow inputs well beyond 512 tokens. We recommend that evaluations include length stress tests and report stability metrics (flip rates, first-flip lengths, entropy trajectories) alongside accuracy.

\paragraph{Defenses and attack cost.}
Truncation to 512 tokens can prevent straightforward Overflip, but is infeasible when full context is required and bypassable by placing malicious content after the first chunk. Compared to gradient-based or prompt-engineered jailbreaks (e.g., DAN, GBDA), Overflip requires no adversarial expertise, only automated repetition, while achieving 8\%--92\% flip rates and substantially higher downstream ASR than padding (41/33 vs.\ 28/6 on GPT-4.1 Mini and Llama 3.1 8B). This combination of low effort, semantic preservation, and black-box applicability makes it a practical threat.

\paragraph{Limitations.}
Our evaluation uses 100 prompts; larger-scale testing may reveal additional edge cases. Positional encoding schemes are inferred from behavior rather than confirmed from documentation. End-to-end reproducibility may be constrained by GPT-4.1 Mini API non-determinism. Defenses are discussed but not exhaustively evaluated; future work should validate mitigations across a broader set of guardrails and deployment scenarios.

%% file: sec/12_conclusion.tex
\section{Conclusion}
\label{sec:conclusion}

We present \emph{Overflip}, a repetition-based attack that exploits a structural vulnerability in lightweight guardrail classifiers: simply repeating a malicious prompt causes the model's prediction to flip from MAL to BEN as the sequence grows beyond the 512-token context boundary. Across 9 widely deployed guardrail models, 5 exhibit such flips at rates of 8\%--92\%, with first flips at 2.6k--9.4k tokens. Mechanistically, repetition homogenizes token-level attention over structural anchor positions, collapsing the model's sensitivity to malicious tokens, a failure mode distinct from traditional attention-dilution baselines. Because the bypassed prompt remains semantically intact, Overflip achieves substantially higher downstream attack success rates than padding (backend ASR: 41 vs.\ 28 on GPT-4.1 Mini; 33 vs.\ 6 on Llama 3.1 8B). These findings expose input length as an under-examined attack surface and highlight positional encoding design as a first-class safety consideration in guardrail development and deployment.

%% file: sec/11_ethics.tex
\section{Ethical Considerations}
\label{sec:ethics}

Overflip could be exploited to bypass content filters in deployed LLM services. We mitigate this risk through responsible disclosure to affected model providers and by planning to release evaluation code and defensive guidance upon completion of institutional IP review. We emphasize that the vulnerability stems from architectural choices (positional bucketing) rather than undisclosed exploits, and that identifying it publicly accelerates the development of more robust guardrails, benefiting the broader safety ecosystem.

\section{Acknowledgements}

LLMs were used for editorial purposes and code generation; all outputs were reviewed by the authors for accuracy and originality, and LLMs are not a core component of our methodology.

%% file: sec/13_appendix.tex
\section{Additional Results}

\begin{figure*}[h]
    \centering
    \includegraphics[width=\textwidth]{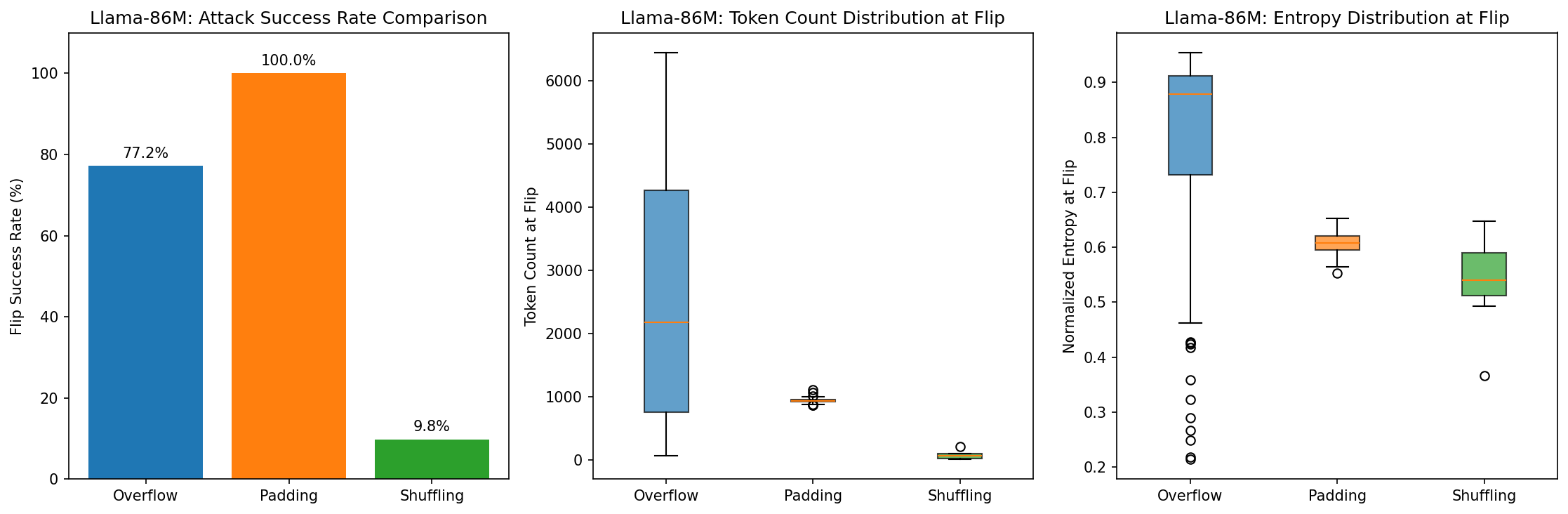}
    \caption{Comparison of three lengthening strategies on the LLaMA Prompt Guard 2 (86M) classifier. \textbf{Left:} MAL$\to$BEN flip success rate for repetition-based Overflip (labeled ``Overflow'' in the plot), benign padding, and sentence shuffling. \textbf{Middle:} distribution of the token count at which the first flip occurs (boxplots over flipped prompts). \textbf{Right:} distribution of normalized CLS-attention entropy at the flip point, where higher values indicate more diffuse attention. Padding achieves the highest flip rate with flips occurring near $\sim$1k tokens, while Overflip flips occur at substantially longer lengths with more variable attention dispersion; shuffling rarely flips.}
    \label{fig:3attack_results}
\end{figure*}

\begin{table*}[h]
\centering
\small
\setlength{\tabcolsep}{4pt}
\begin{tabular}{ccc}
\multicolumn{3}{c}{\textbf{GPT-4.1 Mini}}\\
\begin{tabular}{lrr}
\hline
\multicolumn{3}{c}{overflip ($n=71$)}\\
\hline
Backend LLM  & Count & \% \\
\hline
ac $\rightarrow$ ac & 41 & 57.7\% \\
rej $\rightarrow$ rej & 19 & 26.8\% \\
ac $\rightarrow$ rej & 3 & 4.2\% \\
ac $\rightarrow$ mis & 2 & 2.8\% \\
mis $\rightarrow$ mis & 2 & 2.8\% \\
rej $\rightarrow$ ac & 2 & 2.8\% \\
mis $\rightarrow$ ac & 1 & 1.4\% \\
rej $\rightarrow$ mis & 1 & 1.4\% \\
\hline
\end{tabular}
&
\begin{tabular}{lrr}
\hline
\multicolumn{3}{c}{Padding ($n=92$)}\\
\hline
Backend LLM  & Count & \% \\
\hline
ac $\rightarrow$ ac & 28 & 30.4\% \\
ac $\rightarrow$ mis & 23 & 25.0\% \\
rej $\rightarrow$ rej & 19 & 20.7\% \\
ac $\rightarrow$ rej & 12 & 13.0\% \\
rej $\rightarrow$ mis & 5 & 5.4\% \\
mis $\rightarrow$ mis & 2 & 2.2\% \\
rej $\rightarrow$ ac & 2 & 2.2\% \\
mis $\rightarrow$ rej & 1 & 1.1\% \\
\hline
\end{tabular}
&
\begin{tabular}{lrr}
\hline
\multicolumn{3}{c}{Shuffling ($n=9$)}\\
\hline
Backend LLM  & Count & \% \\
\hline
ac $\rightarrow$ ac & 4 & 44.4\% \\
rej $\rightarrow$ rej & 4 & 44.4\% \\
mis $\rightarrow$ mis & 1 & 11.1\% \\
\hline
\end{tabular}
\\[1.0ex]
\multicolumn{3}{c}{\textbf{Llama 3.1 8B}}\\
\begin{tabular}{lrr}
\hline
\multicolumn{3}{c}{overflip ($n=71$)}\\
\hline
Backend LLM  & Count & \% \\
\hline
rej $\rightarrow$ rej & 33 & 46.5\% \\
ac $\rightarrow$ ac & 13 & 18.3\% \\
ac $\rightarrow$ rej & 7 & 9.9\% \\
mis $\rightarrow$ rej & 5 & 7.0\% \\
ac $\rightarrow$ mis & 3 & 4.2\% \\
mis $\rightarrow$ mis & 3 & 4.2\% \\
rej $\rightarrow$ ac & 3 & 4.2\% \\
mis $\rightarrow$ ac & 2 & 2.8\% \\
rej $\rightarrow$ mis & 2 & 2.8\% \\
\hline
\end{tabular}
&
\begin{tabular}{lrr}
\hline
\multicolumn{3}{c}{Padding ($n=92$)}\\
\hline
Backend LLM  & Count & \% \\
\hline
rej $\rightarrow$ rej & 43 & 46.7\% \\
ac $\rightarrow$ mis & 12 & 13.0\% \\
rej $\rightarrow$ mis & 11 & 12.0\% \\
ac $\rightarrow$ rej & 8 & 8.7\% \\
mis $\rightarrow$ mis & 7 & 7.6\% \\
ac $\rightarrow$ ac & 6 & 6.5\% \\
mis $\rightarrow$ rej & 3 & 3.3\% \\
rej $\rightarrow$ ac & 2 & 2.2\% \\
\hline
\end{tabular}
&
\begin{tabular}{lrr}
\hline
\multicolumn{3}{c}{Shuffling ($n=9$)}\\
\hline
Backend LLM  & Count & \% \\
\hline
rej $\rightarrow$ rej & 3 & 33.3\% \\
ac $\rightarrow$ rej & 2 & 22.2\% \\
rej $\rightarrow$ mis & 2 & 22.2\% \\
ac $\rightarrow$ mis & 1 & 11.1\% \\
mis $\rightarrow$ mis & 1 & 11.1\% \\
\hline
\end{tabular}
\\
\end{tabular}
\caption{Outcome transition counts (Backend LLM ) under three perturbations. Each block sums to 100\% within a perturbation (overflip/Padding/Shuffling). For each type of attack, we observe three types of behaviors: acceptance (\textit{ac}), rejection (\textit{rej}), or misinterpretation (\textit{mis}) the instructions in prompts. Note that the behavior of accaptance here does not necessarily mean that model execute the instructions explicitly, but mean that model implicitly accept the prompt and execute the part of the instructions. From this table, we can find that: overflip attack is the most effective one to retain the semantics; padding attack is the one that causes "misinterpretation" responses in both models to the most. }
\label{tab:orig-to-attacked-transitions-completed}
\end{table*}

%% file: main.bbl
\begin{thebibliography}{32}
\providecommand{\natexlab}[1]{#1}
\providecommand{\url}[1]{\texttt{#1}}
\expandafter\ifx\csname urlstyle\endcsname\relax
  \providecommand{\doi}[1]{doi: #1}\else
  \providecommand{\doi}{doi: \begingroup \urlstyle{rm}\Url}\fi

\bibitem[Anil et~al.(2024)Anil, Durmus, Rimsky, Sharma, Benton, Kundu, Batson,
  Tong, Mu, Ford, Mosconi, Agrawal, Schaeffer, Bashkansky, Svenningsen,
  Lambert, Radhakrishnan, Denison, Hubinger, Bai, et~al.]{anil2024manyshot}
Cem Anil, Esin Durmus, Nina Rimsky, Mrinank Sharma, Joe Benton, Sandipan Kundu,
  Joshua Batson, Meg Tong, Jesse Mu, Daniel~J. Ford, Francesco Mosconi,
  Rajashree Agrawal, Rylan Schaeffer, Naomi Bashkansky, Samuel Svenningsen,
  Mike Lambert, Ansh Radhakrishnan, Carson Denison, Evan~J. Hubinger, Yuntao
  Bai, et~al.
\newblock Many-shot jailbreaking.
\newblock In \emph{Advances in Neural Information Processing Systems
  (NeurIPS)}, 2024.
\newblock URL \url{https://openreview.net/forum?id=cw5mgd71jW}.

\bibitem[Bai et~al.(2023)Bai, Lv, Zhang, Lyu, Tang, Huang, Du, Liu, Zeng, Hou,
  Dong, Tang, and Li]{bai2023longbench}
Yushi Bai, Xin Lv, Jiajie Zhang, Hongchang Lyu, Jiankai Tang, Zhidian Huang,
  Zhengxiao Du, Xiao Liu, Aohan Zeng, Lei Hou, Yuxiao Dong, Jie Tang, and
  Juanzi Li.
\newblock {LongBench: A Bilingual, Multitask Benchmark for Long Context
  Understanding}.
\newblock \emph{arXiv preprint arXiv:2308.14508}, 2023.
\newblock URL \url{https://arxiv.org/abs/2308.14508}.

\bibitem[Chao et~al.(2024)Chao, Debenedetti, Robey, Andriushchenko, Croce,
  Sehwag, Dobriban, Flammarion, Pappas, Tram{\`e}r, Hassani, and
  Wong]{Chao2024JailbreakBenchAO}
Patrick Chao, Edoardo Debenedetti, Alexander Robey, Maksym Andriushchenko,
  Francesco Croce, Vikash Sehwag, Edgar Dobriban, Nicolas Flammarion, George~J.
  Pappas, Florian Tram{\`e}r, Hamed Hassani, and Eric Wong.
\newblock Jailbreakbench: An open robustness benchmark for jailbreaking large
  language models.
\newblock \emph{ArXiv}, abs/2404.01318, 2024.
\newblock URL \url{https://api.semanticscholar.org/CorpusID:268857237}.

\bibitem[{DAIR.AI}(2025)]{PromptingGuide_Adversarial_Jailbreaking_2025}
{DAIR.AI}.
\newblock Adversarial prompting in llms: Jailbreaking, 2025.
\newblock URL
  \url{https://www.promptingguide.ai/risks/adversarial.en#jailbreaking}.
\newblock Accessed: 2025-12-30.

\bibitem[He et~al.(2020)He, Liu, Gao, and Chen]{He2020DeBERTaDB}
Pengcheng He, Xiaodong Liu, Jianfeng Gao, and Weizhu Chen.
\newblock Deberta: Decoding-enhanced bert with disentangled attention.
\newblock \emph{ArXiv}, abs/2006.03654, 2020.
\newblock URL \url{https://api.semanticscholar.org/CorpusID:219531210}.

\bibitem[Huang and Xu(2020)]{huang2020better}
Zhiheng Huang and Peng Xu.
\newblock {Improve Transformer Models with Better Relative Position
  Embeddings}.
\newblock \emph{arXiv preprint arXiv:2009.13658}, 2020.
\newblock URL \url{https://arxiv.org/abs/2009.13658}.

\bibitem[{IBM Research}(2024{\natexlab{a}})]{ibm_2024_granite_guardian_125m}
{IBM Research}.
\newblock {Granite-Guardian-HAP-125m} toxicity classifier, 2024{\natexlab{a}}.
\newblock URL
  \url{https://huggingface.co/ibm-granite/granite-guardian-hap-125m}.
\newblock Release Date: September 6th, 2024.

\bibitem[{IBM Research}(2024{\natexlab{b}})]{ibm_2024_granite_guardian_38m}
{IBM Research}.
\newblock {Granite-Guardian-HAP-38m} lightweight toxicity classifier,
  2024{\natexlab{b}}.
\newblock URL
  \url{https://huggingface.co/ibm-granite/granite-guardian-hap-38m}.
\newblock Release Date: September 6th, 2024.

\bibitem[Ivry and Nahum(2025)]{ivry2025sentinel}
Dror Ivry and Oran Nahum.
\newblock Sentinel: Sota model to protect against prompt injections, 2025.

\bibitem[Jindal(2024{\natexlab{a}})]{jailbreak-detector-2024}
Madhur Jindal.
\newblock Jailbreak detector: Advanced ai security model, 2024{\natexlab{a}}.
\newblock URL \url{https://huggingface.co/madhurjindal/Jailbreak-Detector}.

\bibitem[Jindal(2024{\natexlab{b}})]{jailbreak-detector-large-2024}
Madhur Jindal.
\newblock Jailbreak detector large: Advanced ai security model,
  2024{\natexlab{b}}.
\newblock URL
  \url{https://huggingface.co/madhurjindal/Jailbreak-Detector-Large}.

\bibitem[Jindal(2025)]{Jailbreak-Detector-2-xl-2025}
Madhur Jindal.
\newblock Jailbreak-detector-2-xl: Qwen2.5 chat adapter for ai security, 2025.
\newblock URL
  \url{https://huggingface.co/madhurjindal/Jailbreak-Detector-2-XL}.

\bibitem[Li et~al.(2024)Li, Peng, He, and Yan]{li2024ifrobust}
Zekun Li, Baolin Peng, Pengcheng He, and Xifeng Yan.
\newblock Evaluating the instruction-following robustness of large language
  models to prompt injection.
\newblock In \emph{Proceedings of EMNLP}, 2024.
\newblock URL \url{https://arxiv.org/abs/2308.10819}.

\bibitem[Liu et~al.(2023)Liu, Lin, Hewitt, Paranjape, Bevilacqua, Petroni, and
  Liang]{liu2023lost}
Nelson~F Liu, Kevin Lin, John Hewitt, Ashwin Paranjape, Michele Bevilacqua,
  Fabio Petroni, and Percy Liang.
\newblock Lost in the middle: How language models use long contexts.
\newblock \emph{arXiv preprint arXiv:2307.03172}, 2023.

\bibitem[Liu et~al.(2024)Liu, Jia, Geng, Jia, and Gong]{liu2024promptinjbench}
Yupei Liu, Yuqi Jia, Runpeng Geng, Jinyuan Jia, and Neil~Zhenqiang Gong.
\newblock Formalizing and benchmarking prompt injection attacks and defenses.
\newblock In \emph{USENIX Security Symposium}, 2024.
\newblock URL \url{https://arxiv.org/abs/2310.12815}.

\bibitem[{Meta}(2024)]{meta_2024_llama_prompt_guard_2_22m}
{Meta}.
\newblock {Llama-Prompt-Guard-2-22M} classifier for prompt attacks, 2024.
\newblock URL \url{https://huggingface.co/meta-llama/Llama-Prompt-Guard-2-22M}.

\bibitem[{Meta}(2025)]{meta_2024_llama_prompt_guard_2_86m}
{Meta}.
\newblock {Llama-Prompt-Guard-2-86M} classifier for prompt attacks, 2025.
\newblock URL \url{https://huggingface.co/meta-llama/Llama-Prompt-Guard-2-86M}.

\bibitem[Paulsen(2025)]{Paulsen2025ContextIW}
Norman Paulsen.
\newblock Context is what you need: The maximum effective context window for
  real world limits of llms.
\newblock \emph{ArXiv}, abs/2509.21361, 2025.
\newblock URL \url{https://api.semanticscholar.org/CorpusID:281659451}.

\bibitem[Press et~al.(2022)Press, Smith, and Lewis]{press2022alibi}
Ofir Press, Noah~A Smith, and Mike Lewis.
\newblock Train short, test long: Attention with linear biases enables input
  length extrapolation.
\newblock In \emph{ICLR}, 2022.

\bibitem[ProtectAI.com(2023)]{deberta-v3-base-prompt-injection}
ProtectAI.com.
\newblock Fine-tuned deberta-v3 for prompt injection detection, 2023.
\newblock URL
  \url{https://huggingface.co/ProtectAI/deberta-v3-base-prompt-injection}.

\bibitem[Raffel et~al.(2020{\natexlab{a}})Raffel, Shazeer, Roberts, Lee,
  Narang, Matena, Zhou, Li, and Liu]{raffel2020t5}
Colin Raffel, Noam Shazeer, Adam Roberts, Katherine Lee, Sharan Narang, Michael
  Matena, Yanqi Zhou, Wei Li, and Peter~J Liu.
\newblock Exploring the limits of transfer learning with a unified text-to-text
  transformer.
\newblock \emph{Journal of Machine Learning Research}, 21\penalty0
  (140):\penalty0 1--67, 2020{\natexlab{a}}.

\bibitem[Raffel et~al.(2020{\natexlab{b}})]{Alpher04}
Colin Raffel et~al.
\newblock Exploring the limits of transfer learning with a unified text-to-text
  transformer.
\newblock \emph{Journal of Machine Learning Research}, 21, 2020{\natexlab{b}}.

\bibitem[Shaw et~al.(2018{\natexlab{a}})Shaw, Uszkoreit, and Vaswani]{Alpher03}
Peter Shaw, Jakob Uszkoreit, and Ashish Vaswani.
\newblock Self-attention with relative position representations.
\newblock \emph{NAACL}, 2018{\natexlab{a}}.

\bibitem[Shaw et~al.(2018{\natexlab{b}})Shaw, Uszkoreit, and
  Vaswani]{shaw2018self}
Peter Shaw, Jakob Uszkoreit, and Ashish Vaswani.
\newblock Self-attention with relative position representations.
\newblock In \emph{NAACL}, 2018{\natexlab{b}}.

\bibitem[Sobolik and George(2025)]{sobolik2025llmguardrails}
Tom Sobolik and Vijay George.
\newblock Llm guardrails: Best practices for deploying llm apps securely.
\newblock \url{https://www.datadoghq.com/blog/llm-guardrails-best-practices/},
  October 2025.
\newblock Published October 22, 2025. Accessed May 5, 2026.

\bibitem[Su et~al.(2021)Su, Lu, Pan, Murtadha, Wen, and Liu]{su2021roformer}
Jianlin Su, Yu~Lu, Shengfeng Pan, Ahmed Murtadha, Bo~Wen, and Yunfeng Liu.
\newblock Roformer: Enhanced transformer with rotary position embedding.
\newblock In \emph{arXiv preprint arXiv:2104.09864}, 2021.

\bibitem[Vaswani et~al.(2017)Vaswani, Shazeer, Parmar, Uszkoreit, Jones, Gomez,
  Kaiser, and Polosukhin]{vaswani2017attention}
Ashish Vaswani, Noam Shazeer, Niki Parmar, Jakob Uszkoreit, Llion Jones,
  Aidan~N Gomez, {\L}ukasz Kaiser, and Illia Polosukhin.
\newblock Attention is all you need.
\newblock In \emph{NIPS}, 2017.

\bibitem[Wallace et~al.(2024)Wallace, Xiao, Leike, Weng, Heidecke, and
  Beutel]{wallace2024hierarchy}
Eric Wallace, Kai Xiao, Reimar Leike, Lilian Weng, Johannes Heidecke, and Alex
  Beutel.
\newblock The instruction hierarchy: Training llms to prioritize privileged
  instructions.
\newblock \emph{arXiv preprint arXiv:2404.13208}, 2024.
\newblock URL \url{https://arxiv.org/abs/2404.13208}.

\bibitem[Wang et~al.(2025)Wang, Ji, Wang, Li, Wu, and
  Wang]{SoKJailbreakGuardrails}
Xunguang Wang, Zhenlan Ji, Wenxuan Wang, Zongjie Li, Daoyuan Wu, and Shuai
  Wang.
\newblock Sok: Evaluating jailbreak guardrails for large language models.
\newblock \emph{arXiv preprint arXiv:2506.10597}, 2025.
\newblock URL \url{https://arxiv.org/abs/2506.10597}.

\bibitem[Warner et~al.(2024)Warner, Chaffin, Clavié, and
  et~al.]{ModernBERT_Paper}
Benjamin Warner, Antoine Chaffin, Benjamin Clavié, and et~al.
\newblock Smarter, better, faster, longer: A modern bidirectional encoder for
  fast, memory efficient, and long context finetuning and inference.
\newblock arXiv preprint arXiv:2412.13663, 2024.
\newblock URL \url{https://arxiv.org/abs/2412.13663}.

\bibitem[Yi et~al.(2023)Yi, Xie, Zhu, Kiciman, Sun, Xie, and Wu]{yi2023bipia}
Jingwei Yi, Yueqi Xie, Bin Zhu, Emre Kiciman, Guangzhong Sun, Xing Xie, and
  Fangzhao Wu.
\newblock Benchmarking and defending against indirect prompt injection attacks
  on large language models.
\newblock \emph{arXiv preprint arXiv:2312.14197}, 2023.
\newblock URL \url{https://arxiv.org/abs/2312.14197}.

\bibitem[Zheng et~al.(2024)Zheng, Rana, and Stolcke]{LightweightGuardrails}
Aaron Zheng, Mansi Rana, and Andreas Stolcke.
\newblock Lightweight safety guardrails using fine-tuned bert embeddings.
\newblock \emph{arXiv preprint arXiv:2411.14398}, 2024.
\newblock URL \url{https://arxiv.org/abs/2411.14398}.

\end{thebibliography}
